\documentclass[sigconf]{acmart}

\AtBeginDocument{%
  \providecommand\BibTeX{{%
    \normalfont B\kern-0.5em{\scshape i\kern-0.25em b}\kern-0.8em\TeX}}}

\setcopyright{acmcopyright}
\copyrightyear{2020}
\acmYear{2020}
\acmDOI{10.1145/xxxxxxx.1122456}

\acmConference[Xxxx '20]{Xxxx '20: ACM Symposium on XXX}{June xx--xx, 2020}{Xxxx, XX}
\acmBooktitle{xxx '20: ACM Symposium on XXX,
  June xx--xx, 2020, Xxxx, XX}
\acmPrice{15.00}
\acmISBN{978-1-4503-XXXX-X/18/06}

\usepackage{epsfig}
\usepackage{graphicx}
\usepackage{amsmath}
\usepackage{algorithm}
\usepackage{algorithmic}
\usepackage{subfigure}
\usepackage{multirow}

\newcommand{\eg}{\textit{e.g. }}

\newcommand{\ie}{\textit{i.e. }}

\begin{document}

\title{Object-QA: Towards High Reliable Object Quality Assessment}

\author{Jing Lu$^{1*}$, Baorui Zou$^{2*}$, Zhanzhan Cheng$^{31*}$, Shiliang Pu$^1$, Shuigeng Zhou$^2$, Yi Niu$^{1}$, Fei Wu$^3$}
\thanks{\textsuperscript{*}Authors contributed equally to this research.}
\affiliation{%
  \institution{\textsuperscript{1}Hikvision Research Institute, China;~~\textsuperscript{2}Fudan University, Shanghai, China;~~\textsuperscript{3}Zhejiang University, Hangzhou, China}
}
\begin{abstract}
In object recognition applications, object images usually appear with different quality levels.
Practically, it is very important to indicate object image qualities for better application performance, \eg filtering out low-quality object image frames to maintain robust video object recognition results and speed up inference.
However, no previous works are explicitly proposed for addressing the problem.
In this paper, we define the problem of object quality assessment for the first time and propose an effective approach named Object-QA to assess high-reliable quality scores for object images.
Concretely, 
Object-QA first employs a well-designed relative quality assessing module that learns the intra-class-level quality scores by referring to the difference between object images and their estimated templates.
Then an absolute quality assessing module is designed to generate the final quality scores by aligning the quality score distributions in inter-class.
Besides, Object-QA can be implemented with only object-level annotations, and is also easily deployed to a variety of object recognition tasks.
To our best knowledge this is the first work to put forward the definition of this problem and conduct quantitative evaluations.
Validations on 5 different datasets show that Object-QA can not only assess high-reliable quality scores according with human cognition, but also improve application performance.
\end{abstract} 

\maketitle

\section{Introduction}
\label{intro}

\begin{figure}
    \begin{center}
    \includegraphics[width=0.99\linewidth]{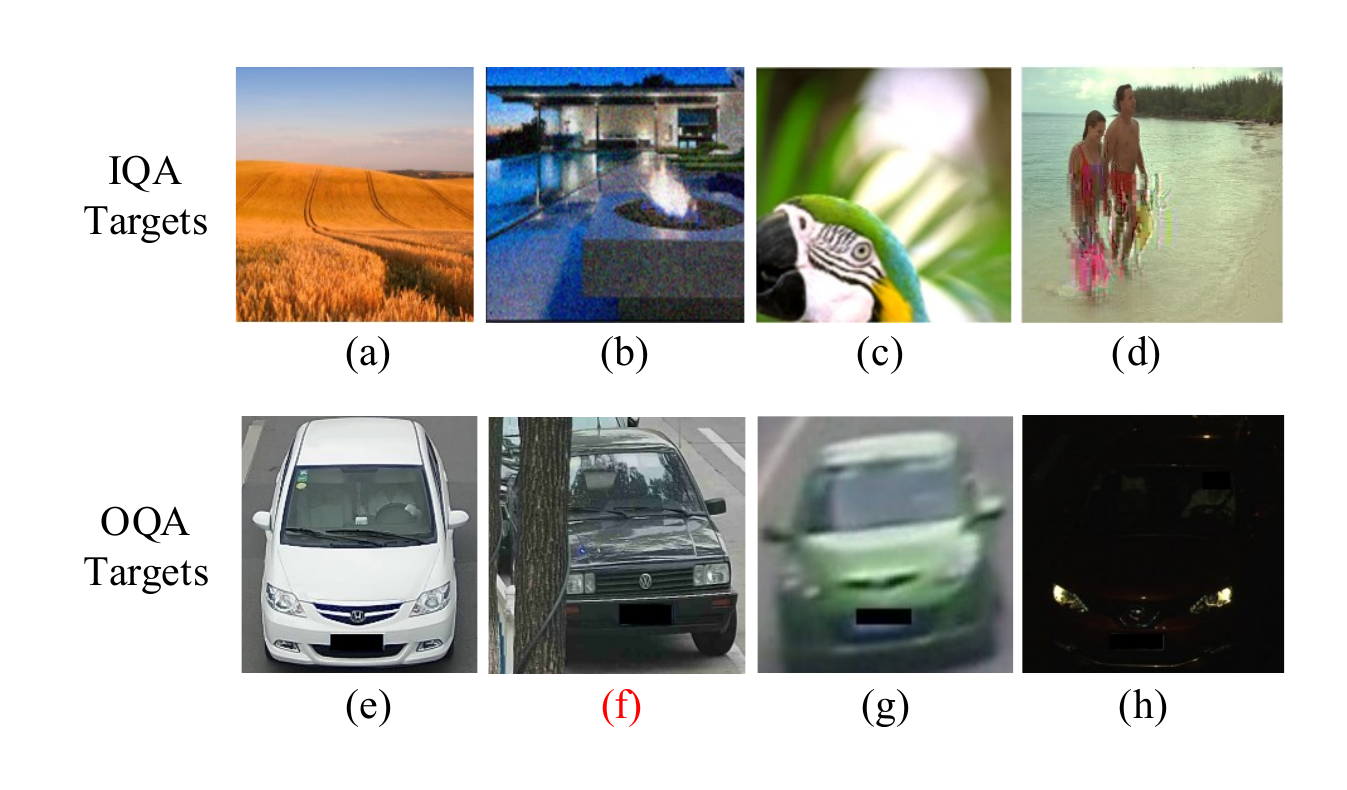}
    \end{center}
    \vspace{-0.5cm}
    \caption{The comparison of assessing target between IQA and OQA.
    (a)$\sim$(d) are examples of IQA: (a) high-quality image containing no object, (b) low-quality image with gaussian noise containing no object, (c) low-quality image containing 1 clear object but with blurred background, (d) low-quality image with 2 distorted objects.
    (e)$\sim$(h) are examples of OQA: (e) high-quality image with 1 vehicle, (f) low-quality image with 1 clear but occluded vehicle (IQA would regard it as a \textbf{high}-quality image since no physical noise occurs, which is a failed case of IQA methods applied on object images), (g) low-quality image with 1 blurred vehicle, (h) low-quality image with 1 vehicle under poor illumination.
    }
    \vspace{-0.5cm}
    \label{fig-IQAvsOQA}
\end{figure}

\begin{table*}
\begin{center}
\scalebox{1}{
\begin{tabular}{ccc}
\hline
\multicolumn{1}{c}{Problem} & \multicolumn{1}{c}{IQA} & \multicolumn{1}{c}{OQA} \\
\hline
\hline
\multicolumn{1}{c|}{assessing target} & \multicolumn{1}{c|}{whole image} & cropped object image\\
\hline
\multicolumn{1}{c|}{object number} & \multicolumn{1}{c|}{none/multiple} & one\\
\hline
\multicolumn{1}{c|}{background quality} & \multicolumn{1}{c|}{consider} & ignore \\
\hline
\multicolumn{1}{c|}{\multirow{2}{*}{noise type}} & \multicolumn{1}{c|}{constrained physical noise (\eg} & {unconstrained noise (\eg} \\
\multicolumn{1}{c|}{} & \multicolumn{1}{c|}{gaussian blur/noise, transmission errors} & {occlusion, motion blur, poor illumination} \\ 
\hline
\multicolumn{1}{c|}{\multirow{2}{*}{application}} & \multicolumn{1}{c|}{image processing task (\eg} & {object recognition task (\eg} \\
\multicolumn{1}{c|}{} & \multicolumn{1}{c|}{image restoration, image super-resolution} & {person/vehicle re-identification, text recognition} \\
\hline
\end{tabular}
}
\end{center}
\caption{Differences between OQA and IQA.
}
\label{tab_oqa_vs_iqa}
\end{table*}

\begin{figure*}
    \begin{center}
    \includegraphics[width=0.9\linewidth]{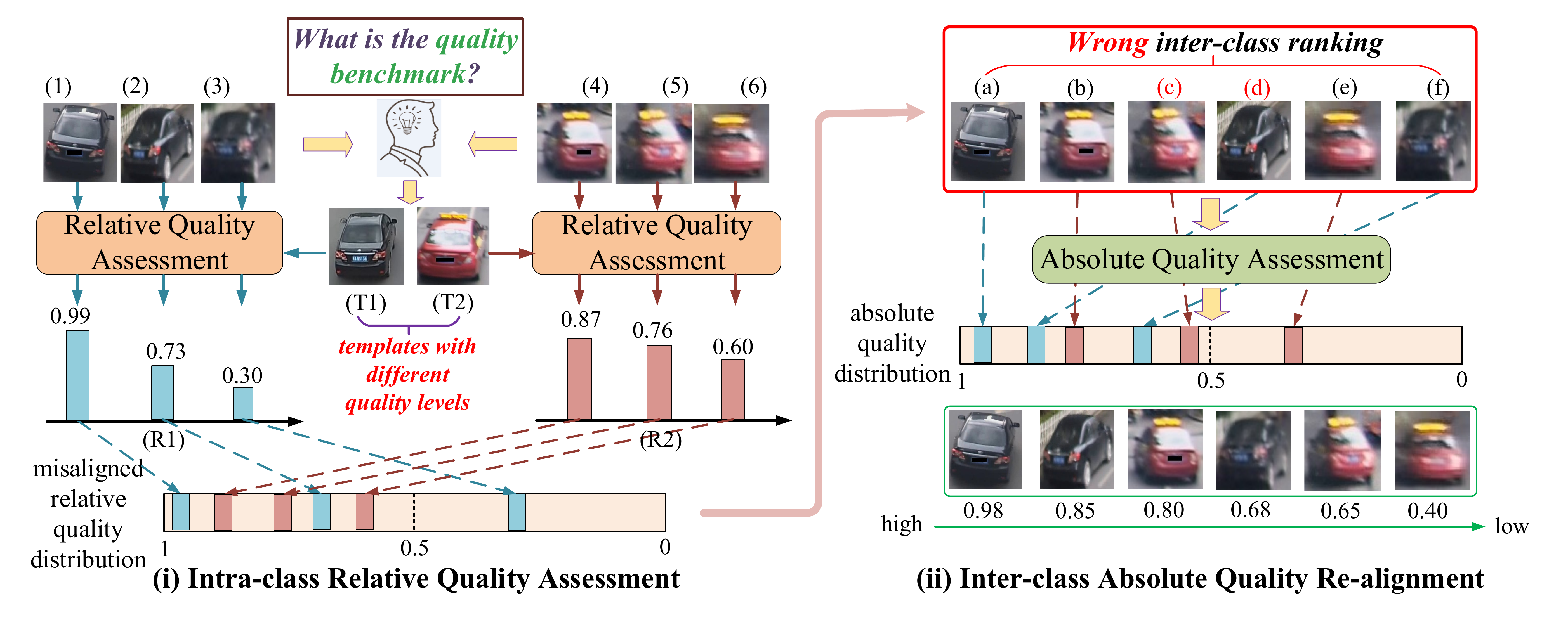} 
    \end{center}
    \vspace{-0.8cm}
    \caption{Illustration of object quality assessment.
    }
    \label{fig-motivation}
\end{figure*}

In many real-life object recognition systems (\eg vehicle \cite{Wang2019ICCV} or person \cite{Li2018CVPR} re-identification, license plate recognition \cite{Xu2018Towards} and scene text recognition \cite{Bai2018CVPR}), object images usually contain one centering object and appear with different qualities (as (e)$\sim$(h) in Figure \ref{fig-IQAvsOQA}) caused by complicated environmental inteferences (\eg occlusion, motion, illumination change).
Practically, high-quality (\eg clear, undistorted) object images are beneficial to recognition, while low-quality (\eg blurred, distorted) object images may result in poor recognition performance.

Generally, high reliable quality scores of object images can be used to enhance application performance.
For instance, (1) object quality assessment can be used as a pre-processing technique: filtering out low-quality images to avoid wrong results, boosting applications' efficiency and even saving storage space;
(2) It can also be applied in post-processing: removing unreliable recognitions caused by low quality to promote application experience.
Thus, \emph{identifying object image quality} can be a powerful strategy to maintain robust application performance.


\textbf{Problem definition.}
Here, we first explicitly present the definition of \emph{\textbf{o}bject \textbf{q}uality \textbf{a}ssessment} (\emph{abbr.} OQA) problem: given a specific recognition/classification task in which each image normally only contains one object for recognition ((e)$\sim$(h) in Figure \ref{fig-IQAvsOQA}), the quality level of each object image requires to be assessed with reasonable quality scores by considering how well the object image is beneficial for recognition (\ie higher scores represent better quality and vice versa), in this way the quality relations among any object images in this task can be determined.

An intuitive application of OQA is to filter out relative low-quality images tending to result in wrong recognitions, so that recognition/classification tasks can be enhanced.
Emphatically, the object quality scores are domain-specific. For example, the quality scores can reasonably express the quality relations among different vehicle images, but can not distinguish the quality between a vehicle image and a person image.

The OQA task is 
a different research topic compared with the well-know \textbf{i}mage \textbf{q}uality \textbf{a}ssessment (\emph{abbr.} IQA) in the following aspects: (1) assessing target: IQA attempts to assess the quality of the entire image in which none or multiple objects may exist, and jointly considers the quality of background area and foreground objects (see (a)$\sim$(d) in Figure \ref{fig-IQAvsOQA}), while OQA simply evaluates the quality of the only foreground object in the cropped image (see (e)$\sim$(h)); (2) goal: IQA predicts the distortion extent of the entire image by considering constrained types of physical noises (\eg gaussian blur/noise, JPEG2000 compression and fast fading channel distortion) \cite{Sheikh2006statisti}, thus it can be applied in image restoration \cite{KatsaggelosDigital}, image super-resolution \cite{OuwerkerkImage} and even image retrieval \cite{Yan} applications. But the goal of OQA is to achieve object-level quality estimation indicating whether object images are suitable for recognition, by considering unconstrained noises (\eg occlusion, blur and poor illumination) in real-life scenarios, so it can be used for boosting the performance and speed of various object recognition systems.
Figure \ref{fig-IQAvsOQA}(f) is an object image with no physical noise but containing occluded object, IQA may generate high scores while OQA should consider it a low-quality image since its recognition result is very possibly incorrect. We summarize the differences between OQA and IQA in Table \ref{tab_oqa_vs_iqa}.

\textbf{Proposed idea.}
In this paper, our focus is the OQA task. Firstly we consider an intriguing question that how exactly does human identify object's quality level?

As illustrated in many works from IQA, human visual system needs a reference image to quantify the perceptual discrepancy by comparing the distorted image either directly with the original undistorted image (reference based methods \cite{SeoDeep, Prashnani2018CVPR}) or implicitly with a hallucinated image in mind (no-reference based methods \cite{LinHallucinated, hongyu2018AAAI}). 
We transfer this common knowledge to OQA, that is,

given an object image, 
human may firstly assume the high-quality morphology of this object class as its quality benchmark, like the benchmark (T1) for object images (1)$\sim$(3) in Figure \ref{fig-motivation},
then estimate the difference between the object image and its benchmark as its quality.
Smaller difference represents better quality and vice versa.
We describe the quality benchmark as `template' and denote this class-aware template comparison strategy as \emph{relative quality assessment}.
As shown in (R1)/(R2) of part (i), the predicted quality scores can reasonably express the relative intra-class quality relations.

However, inter-class quality score distribution is still unreasonable (`relative quality distribution' in part (i) leads to the wrong quality ranking of image (c) and (d) in part (ii)).
It's because the hypothetic `template' is subjective and varies from person to person, making the estimated templates of different classes correspond to inconsistent quality levels, as template (T1) and (T2) in part (i) imply,
thus the quality scores between different classes are incomparable directly.
Fortunately, we discover that the misordered ranking can be re-aligned by judging whether object images are beneficial to correct recognition (similar idea appeared in \cite{liu2017quality}). For instance in part (ii), image (d) has a better chance for producing correct recognition result compared with (c) since the contour of vehicle in (d) can still be identified, then the wrong ranking in between can be fixed. Based on this observation, we can re-align the quality relations among different classes while maintaining the 
quality distribution learned by \emph{relative quality assessment}, and obtain rational `absolute quality distribution' (see part (ii)). Such sorting strategy is denoted as \emph{absolute quality assessment} in inter-class.

On the basis of above observations, we propose an effective object-level quality assessment approach named Object-QA to assess high-reliable quality scores for object images.
Specifically, Object-QA consists of two modules: the Relative Quality Assessing module (\emph{abbr.} RQA) for obtaining intra-class-level quality scores, and the Absolute Quality Assessing module (\emph{abbr.} AQA) for obtaining inter-class-level quality scores.
RQA first synthesizes hypothetical templates for each class of object images, then computes the distances between selected images and their templates as their relative quality scores.
AQA is responsible for learning the quality distribution alignment from different classes to obtain the final absolute quality scores in inter-class. 
Main contributions are:

(1) We give the explicit definition of \emph{object quality assessment} problem for the first time.

(2) We propose a general object quality assessment approach named Object-QA for assessing high-reliable quality scores with only recognition-based annotations.
Inspired by human cognition \cite{LinHallucinated,liu2017quality}, we implement Object-QA with two modules: the relative quality assessing module for obtaining intra-class quality scores and the absolute quality assessing module for generating inter-class quality scores.
The proposed network is quite light-weighted and can be easily equipped on various object recognition applications for improving their performance and inference speed, yet brings extremely low computational cost.


(3) Experiments on 5 tasks, including 1 synthetic MNIST dataset and 4 real-world scenarios (person \cite{Taiqing2014Person} and vehicle \cite{Zapletal2016Vehicle} re-identification, video text spotting \cite{karatzas2013icdar}, license plate recognition \cite{Xu2018Towards}) demonstrate
the effectiveness of Object-QA.

\section{Related work}
\label{related_work}

In fact, no previous work can be found for specially study the problem of object quality assessment, which is a stark contrast to its potential power for improving the robustness and time efficiency of object recognition systems. In the few related works, object quality assessment strategies are all existed as by-products in video-based recognition tasks, and these methods didn’t report any quantitative results regarding quality assessment.

Here we categorize them as follows:
(1) \textbf{attention-based strategy.} Most previous methods fall into this category by adopting temporal attention mechanisms to indicate the attending weight of different images
, the attending weights are treated as quality scores.
Liu et al. \cite{liu2017quality} and  Yang et al. \cite{Yang2017Neural} adopted a temporal attention module to aggregate discriminative features from multiple frames for video person re-identification and face recognition respectively, then attention weight of each frame was treated as the quality score.
Song et al. \cite{Song2018Region} also followed this quality assessment routine for person re-identification task.
(2) \textbf{feature clustering strategy.}
Recently Cheng et al. \cite{Cheng2019You} proposed a quality scoring method for selecting the highest-quality text regions from tracked text streams, in which quality scores were generated by computing the feature distance between testing images and their templates obtained by K-means clustering.

All the above strategies only concentrate on depicting the intra-class quality relations between consecutive object frames in video-based tasks, but are limited to assessing quality in more general single-image scenario, in which quality relations between images from different classes needs identifying.

Overcoming the limitation of previous work, in this paper we explicitly explore the quality correlations among object images from different classes, and propose a robust object quality assessment method to generate absolute quality scores for object images with only recognition-based annotations.

\section{Methodology}
\begin{figure*}
\begin{center}
\includegraphics[width=0.9\linewidth]{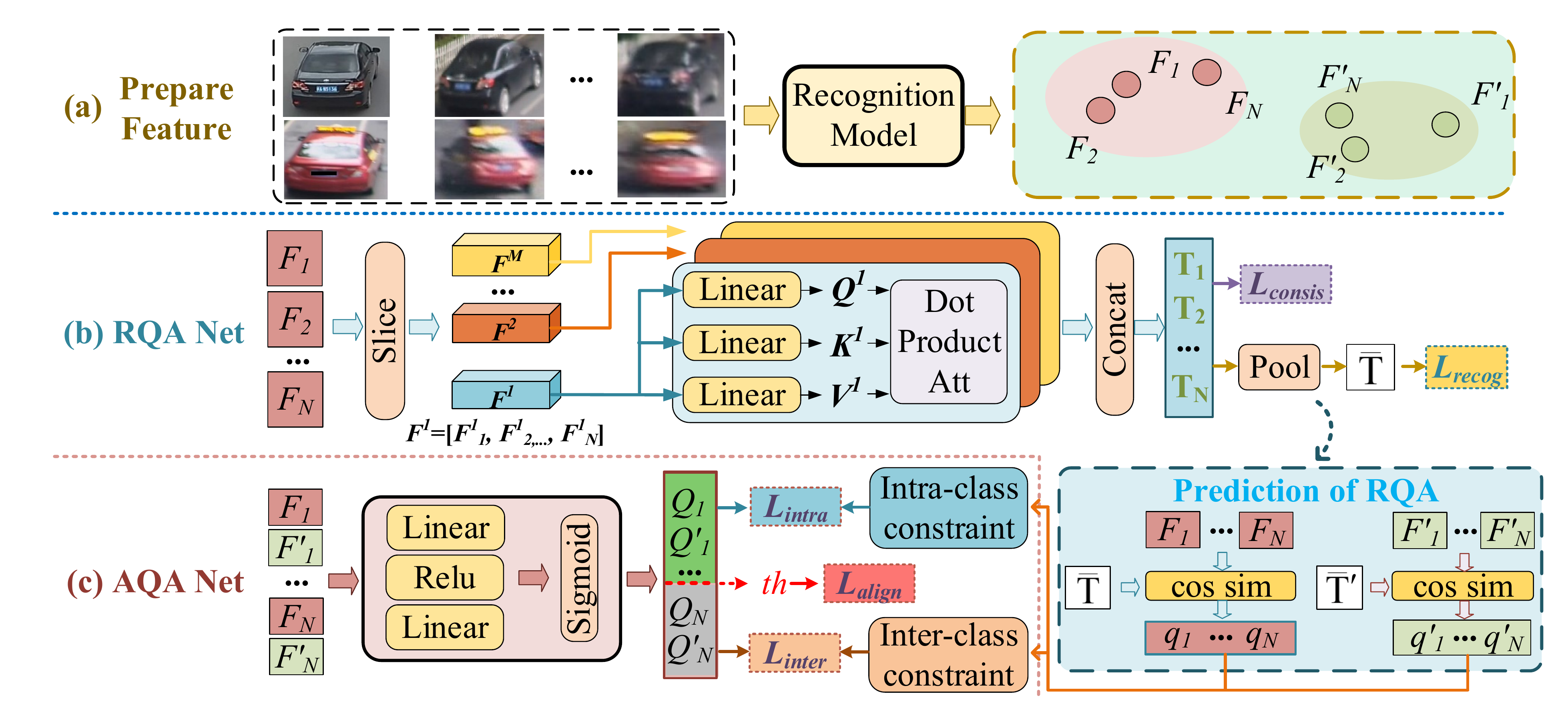}
\end{center}
\vspace{-0.5cm}
   \caption{
    The workflow of Object-QA, which consists of 3 steps:
    (a) Extracting features for RQA and AQA with a pre-trained recognition model;
    (b) Generating relative quality scores by RQA network;
    (c) Generating the final absolute quality scores by AQA network.
    Details of neural networks are described in \emph{Experiment} section.
   }
\label{fig_our_frame}
\end{figure*}

\subsection{Overview}
The workflow of our method is illustrated in Figure \ref{fig_our_frame}, which consists of three steps:
    (1) extracting features for the following parts by a pre-trained recognition model,
    (2) learning the relative quality ranking between intra-class images by Relative Quality Assessment (\emph{abbr}. RQA) network,
    and (3) adjusting the learned relative quality scores by Absolute Quality Assessment (\emph{abbr}. AQA) network to obtain final absolute quality scores.

Specifically, RQA first synthesizes a template (representing the relative quality benchmark) by exploring the quality relations between intra-class object images, then generates relative quality scores by calculating the differences between testing images and their templates.
AQA is responsible for re-aligning relative quality scores among different classes by considering correlations between relative quality distributions, and generating absolute quality scores.

During inference only AQA is needed, which can be easily deployed in existing recognition network as a quality generation branch.

\subsection{Pre-feature Extracting}
Given a set of intra-class object images $I = \{I_{1},I_{2},...,I_{N}\}$, we extract their features as $F = [F_{1},F_{2},...,F_{N}]\in \mathbb{R}^{C\times N}$ with a pre-trained recognition model.
The recognition models adopted in different scenarios are detailed in \emph{Experiment}.

\subsection{Relative Quality Assessment}
RQA is designed to generate relative quality scores, which includes 2 major steps:
1) synthesizing the hypothetical template represented in feature space,
and 2) producing the quality scores by calculating the distance between testing images and their templates.

\emph{Synthesizing template.}
Empirically, the features of different quality images are actually localised at different positions in feature space, which implies the quality correlations are contained in feature distributions intrinsically.
Naturally, the represented template should be closer to those high-quality images with correct recognitions while be farther from the low-quality ones easily leading to wrong recognitions.
Here, we adopt the multi-head attention structure \cite{Vaswani2017Attention} to explore the quality correlations among different quality images in intra-class, then further synthesize the feature representation of template.

Specifically, $F\in \mathbb{R}^{C\times N}$ is firstly sliced into $M$ pieces $(F^1, F^2,..., F^M)$ along the channel dimension ($C$),
and then they are separately fed into M heads.
Concretely, in the $m$-th head, $F^m$ is firstly fed into the linear layer to obtain the triple of query, key and value: $(Q^{m}, K^{m}, V^{m})$.
Then we conduct the scaled dot-product attention \cite{Vaswani2017Attention} on each head to obtain head outputs,
\begin{equation}
{T}^{m} \text{=} softmax(\frac{Q^{m}*V^{m}}{\sqrt{d}})*V^{m},
\label{att_syn_template_1head}
\end{equation}
where `*' is dot-product and $\sqrt{d}$ is the scale factor according to \cite{Vaswani2017Attention}.
Finally, the corresponding template of $F$ is synthesized by directly concatenating all head outputs along the channel dimension, i.e.,
\begin{equation}
{T} \text{=} Concat(T^{1},T^{2},...,T^{M}),
\label{att_syn_template_allhead}
\end{equation}
where $T\in \mathbb{R}^{C\times N}$.
Then the template feature representation $\bar{T}\in \mathbb{R}^{C}$ of each class is generated by conducting average pooling on $T$.

Theoretically, the optimal template feature should be the one that is most beneficial to recognition. So
in order to learn the templates, we use object recognition loss (denoted by $\mathcal{L}_{recog}$) during optimization, so that the estimated template feature can approximate the optimal one. 
Besides, the intra-class quality scores should be computed with respect to the same quality benchmark, thus a template consistency loss (denoted by $\mathcal{L}_{consis}$) is introduced to constrain the template feature of each image in the same class being consistent.
The loss function of RQA is formalized as: 
\begin{equation}
\mathcal{L}_{RQA} \text{=} \mathcal{L}_{recog} + \lambda_{c}\mathcal{L}_{consis}
\label{relative_loss}
\end{equation}
where $\lambda_{c}$ is a tunable parameter and the consistency loss is:
\begin{equation}
\mathcal{L}_{consis} \text{=} \frac{2}{N*(N-1)}\sum_{i=1}^{N-1}\sum_{j=i+1}^{N}MSE(T_{i},T_{j})
\label{relative_consistency_loss}
\end{equation}
in which $MSE$ is the average mean square error. $T_i\in\mathbb{R}^C$ and $T_j\in\mathbb{R}^C$ mean the learned templates of image $F_i$ and $F_j$, respectively.

\emph{Generating quality scores.}
Relative quality score $q_i$ can be generated by computing the feature distance between testing image and its class's template, i.e.,
\begin{equation}
{q}_{i} \text{=} \frac{\bar{T} * F_i}{||\bar{T}||*||F_i||},
\label{relative_score}
\end{equation}

\subsection{Absolute Quality Assessment}
Since the learned templates of different classes usually correspond to different quality levels, RQA is incapable of assigning reasonable quality scores in inter-class.
To address this, AQA network is designed to re-adjust the relative quality scores and produce absolute quality scores, which contains three progressive parts as follows.


\emph{Binary inter-class quality alignment.}
We first align the quality distributions of different classes by referring to a learnable quality anchor $th$.
That is, those images with quality score larger than the quality anchor always can be recognized correctly, vice versa.
Concretely,
we assume that all scores fall into the range (0,1). 
Then all images should be mapped into 2 absolute quality score ranges: the correctly recognized ones are mapped into the higher score range $[th,1)$ while the wrongly recognized ones are mapped into the lower score range $(0,th)$. This alignment is ensured by:
\begin{equation}
\mathcal{L}_{align} \text{=} \frac{1}{M} \sum_{i=1}^{M} relu[\delta(I_i)*(q_{i}-th-\epsilon)]
\label{align_loss}
\end{equation}
where indicator function $\delta(I_i)$ equals -1 when $I_{i}$ can be correctly recognized and 1 otherwise, and $\epsilon$ is a regularization parameter for avoiding regressing the same quality scores. Quality anchor $th$ is jointly optimized with $q_{i}$.


\emph{Intra-class quality distribution maintaining.}
Obviously the learned relative quality ranking in intra-class should be kept.
Given the relative quality score triplet $(q_{1},q_{2},q_{3})$ in a class, the corresponding absolute score triplet generated by AQA are $(Q_{1},Q_{2},Q_{3})$.
The absolute quality ranking of each image pair should remain the same as relative scores, which is constrained in a loss function:
\begin{equation}
\mathcal{L}_{rank} \text{=} \frac{1}{3} \sum_{i=1}^{2}\sum_{j=i+1}^{3}relu[\Delta(q_i,q_j)*(Q_i-Q_j-\epsilon)]
\label{intra-ranking-score}
\end{equation}
where $\Delta(\cdot,\cdot)$ equals -1 when $q_{i}>q_{j}$ and 1 otherwise.

Additionally, the correlation between quality differences $d_{1}=|q_{1}-q_{2}|$ and $d_{2}=|q_{2}-q_{3}|$ ($|\cdot|$ denotes absolute value) should also be taken into consideration for better characterizing the intra-class quality distribution, which is denoted as:
\begin{equation}
\mathcal{L}_{d-rank} \text{=} relu[\Delta(d_1,d_2)*(d_1-d_2-\epsilon)].
\label{intra-d-ranking-score}
\end{equation}
Then total loss function for keeping intra-class quality ranking is:
\begin{equation}
\mathcal{L}_{intra} \text{=} \mathcal{L}_{rank} + \lambda_{intra}\mathcal{L}_{d-rank},
\label{intra-score}
\end{equation}
where $\lambda_{intra}$ is a tunable parameter.

\emph{Inter-class quality distribution maintaining.}
We firstly define the quality distribution entropy ($e$) to depict the quality distribution in each quality score triplet $(s_1, s_2, s_3)$:
\begin{equation}
{e} \text{=} -\frac{min(D_1,D_2)}{max(D_1,D_2)}\log(\frac{min(D_1,D_2)}{max(D_1,D_2)})
\label{quality distribution entropy}
\end{equation}
where $D1=|s_1-s_2|$ and $D2=|s_2-s_3|$.
Larger value of $e$ implies more disordered quality distribution while smaller value represents more uniform distribution.

Given two relative quality score triplets $(q^A_1,q^A_2,q^A_3)$ and $(q^B_1,q^B_2,q^B_3)$ from two different classes $A$ and $B$, we can obtain their relative $e$ as $e^A_r$ and $e^B_r$.
Correspondingly, the absolute $e$ are represented as $e^A_a$ and $e^B_a$.
Then we can align quality distributions between class A and class B by:
\begin{equation}
\mathcal{L}_{inter} \text{=} relu[\Delta(e^A_r,e^B_r)*(e^A_a-e^B_a-\zeta)]
\label{quality distribution entropy}
\end{equation}
where $\zeta$ is a regularization parameter.

Finally, AQA is optimized as:
\begin{equation}
\mathcal{L}_{AQA} \text{=} \mathcal{L}_{align} + \lambda_{a1}\mathcal{L}_{intra} + \lambda_{a2}\mathcal{L}_{inter}
\label{absolute loss}
\end{equation}
where $\lambda_{a1}$ and $\lambda_{a2}$ are tunable parameters. 
\section{Experiments}
Since all the previous works didn’t report any quantitative results regarding object quality assessment, we can only re-implement two representative state-of-art strategies (\cite{liu2017quality}, \cite{Cheng2019You}) for comparison. 
We first explore the effectiveness of our method on a synthetic MNIST \cite{LeCun1998Mnist}, then conduct extensive experiments on 4 different real-life scenarios to demonstrate its robustness and potential power for enhancing recognition performance.

Unfortunately we are unable to compare our method with previous literature in IQA for two reasons: (1) training of IQA methods requires the availability of annotated quality labels or reference images, which is absent in existing object classification/recognition datasets, making the re-implementation impossible;
(2) to say the least, even if we re-implement IQA methods, since IQA only considers constrained physical noise, the performance of IQA methods will degenerate in OQA-oriented scenarios where unconstrained types of interferences exist (\eg occlusion, pose changes), a failed case is shown in Figure \ref{fig-IQAvsOQA}(f).



\subsection{Implementation Details}
Our work is built on the `PyTorch' framework.

\textbf{The pre-trained recognition model}.
In five evaluation scenarios, the pre-trained recognition models mentioned in Section 3.2 are firstly pre-trained, applying the same training strategies as used in \cite{LeCun1998Mnist}, \cite{GaoRevisiting}, \cite{Liu2018PROVID}, \cite{cheng2017focusing} and \cite{shi2016robust}.
Implementation details can be found in our supplementary meterial.

\textbf{Relative quality assessment network}.
In each training batch, RQA takes the features of 3 randomly selected images from the same class as input, followed by a multi-head attention structure.
The num of heads ($M$) is set to 4 in all of our experiments.
In the scaled dot-product attention module, 3 fully connected layers are employed for embedding the input feature and outputting $Q$, $K$, $V$, and the output channel number is the same as input feature.
During training, $\lambda_{c}$ is set to 1 and batch size is set to 9600,
and `Adam' is used for optimization with learning rate = 0.0005 and decay rate = 0.94 for every 200 epochs.

\textbf{Absolute quality assessment network}.
The regularization term $\epsilon$ and $\zeta$ are set to 0.02 and 0.01 respectively. 
$\lambda_{intra}$, $\lambda_{a1}$ and $\lambda{_{a2}}$ in Equation \ref{absolute loss} are all set to 1.
In training stage, batch size is set to 9600, and `Adam' is used to optimize the model with fixed learning rate=$10^{-3}$ and weight decay=0.0005.

\subsection{Evaluation protocol}
The fact is that existing classification/recognition datasets all lack the ground truth of quality scores. Besides, the annotation of object quality is tricky since the accuracy of annotated scores can not be ensured due to the subjective perceptual errors of different human annotators. 
Thus we attempt to design some strategies to obtain the ground truth of quality scores for quantitative evaluations (detailed strategies for ground truth generation are illustrated in corresponding validation scenario).

Two quality evaluation metrics are adopted to evaluate our method:
    1) Spearman’s Rank Order Correlation Coefficient (SROCC) \cite{LinHallucinated} for evaluating monotonic relations between ground truth and predicted scores,
    2) Linear Correlation Coefficient (LCC) \cite{LinHallucinated} for assessing linear correlations between ground truth and predictions.
Additionally, we also validate the improvements on recognition performance as an auxiliary evidence to prove the effectiveness of our method.

\subsection{Ablation Study} 
\textbf{Dataset}.
Arising from the lack of object quality assessment benchmarks, we generate a MNIST \cite{LeCun1998Mnist}-based quality assessment dataset in which images appear in different qualities. Correspondingly, quality score labels are provided for evaluation.

The synthetic MNIST dataset is generated by introducing two quality factors:
1) Blur: gaussian blur with evenly spaced kernel sizes of [3,5,7,9,11,13,15,17,19] for synthesizing images with different blur degrees. Thereby, a total of 9 images are synthesized respecting to 1 original image.
2) Illumination deterioration: contrast and intensity modification for generating images with various illumination conditions, in which contrast changes from 0.4 to 0.9 and intensity changes from -160 to 250.
3) Mixed: considering both blur and illumination.

The object quality scores should be able to represent the extent of how an object image is suitable for recognition. Consider that the original MNIST image is noise-free, so it can be regarded as the `ideal' template which is most beneficial for recognition. Then it's reasonable to generate the ground truth of quality score by calculating feature distances between the original MNIST images and their polluted images.
Note that, the ground truth is not used in training but only used to evaluate the method's performance.

We randomly select half of the images in each class as the training set, i.e. 600 thousand images with blur and 400 thousand images with illumination, while the rest of them are used for testing.
In order to evaluate our method, we design two test sets: the `Intra' and `Inter' set.
`Intra' set contains 10000 groups of images, in which images in a group are from the same class.
`Inter' set contains 10000 groups of images, in which images in a group are from different classes.
The group size varies from 3 to 10.
The pre-trained model is the same to \cite{LeCun1998Mnist}, and the outputs before the final classification layer are used as the extracted features.

\textbf{Evaluation protocol}.
We first compute SROCC and LCC in each image group, then report the average value of all groups. 

\begin{table}
\begin{center}
\scalebox{1}{ 
\begin{tabular}{cccccccc}
\hline
\multicolumn{2}{c}{\multirow{2}{*}{Metrics}} & \multicolumn{2}{c}{Blur} & \multicolumn{2}{c}{Illumination} & \multicolumn{2}{c}{Mixed} \\
\cline{3-8}
\multicolumn{2}{c}{} & RQA & AQA & RQA & AQA & RQA & AQA \\
\hline
\multirow{2}{*}{SROCC} & Intra & 0.874 & 0.857 & 0.933 & 0.929 & 0.920 & 0.893 \\
\multirow{2}{*}{} & Inter & 0.641 & 0.776 & 0.765 & 0.805 & 0.767 & 0.814 \\
\hline
\multirow{2}{*}{LCC} & Intra & 0.880 & 0.861 & 0.966 & 0.950 & 0.957 & 0.936\\
\multirow{2}{*}{} & Inter & 0.680 & 0.866 & 0.850 & 0.886 & 0.845 & 0.895 \\
\hline
\end{tabular}
}
\end{center}
\caption{Quality assessment evaluation of RQA and AQA under different conditions.
`Blur' and `Illumination' mean images are only polluted by blur and lighting, respectively.
`Intra' and `Inter' denote evaluations on `Intra' and `Inter' sets respectively. 
}
\vspace{-0.5cm}
\label{tab_mnist}
\end{table}
%

\begin{figure}
\begin{center}
\subfigure[MNIST]{
\label{fig_score_curve_mnist}
\includegraphics[width=0.8\linewidth]{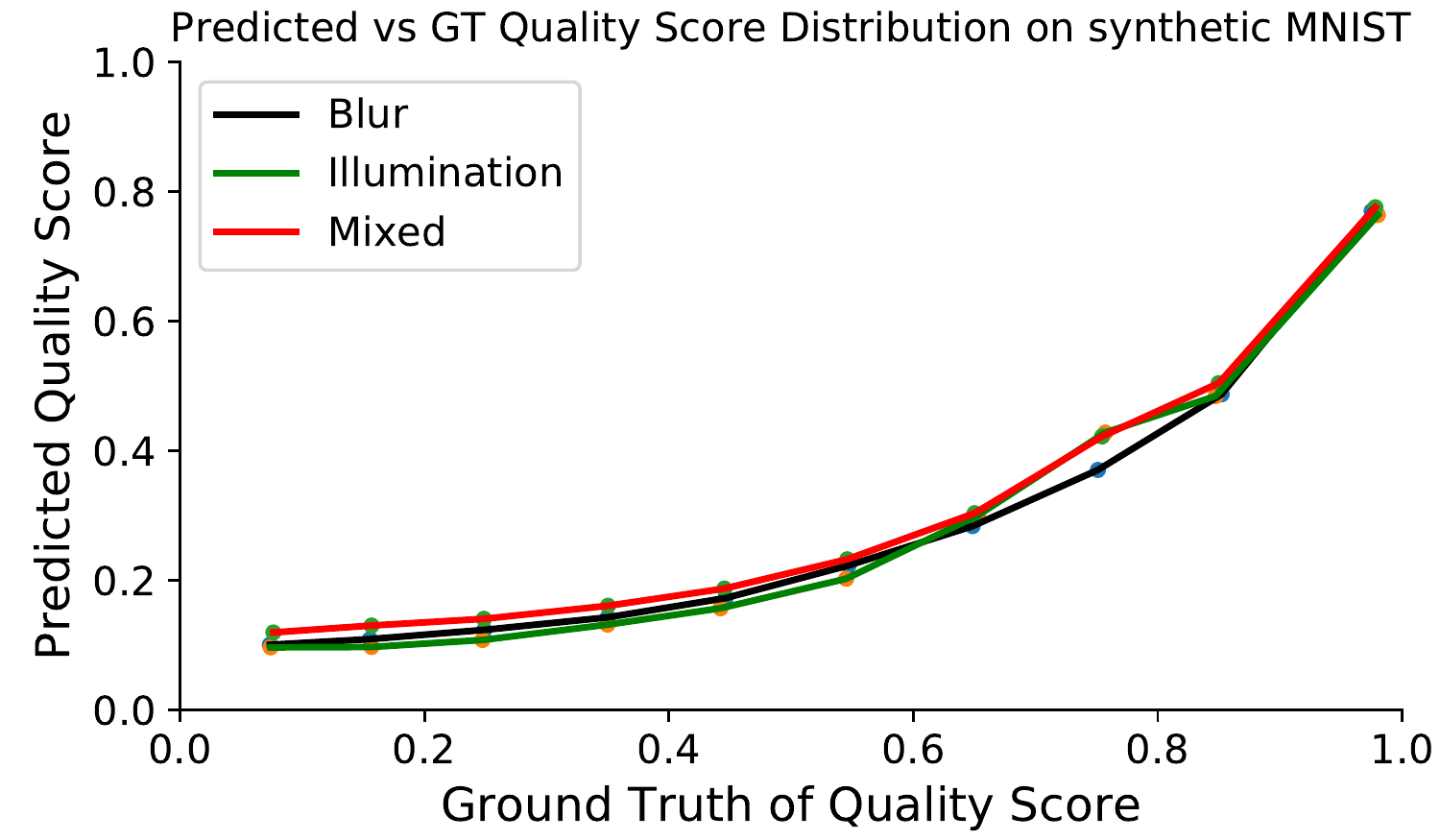}}
\subfigure[CCPD]{
\label{fig_ccpd_qscore_distri}
\includegraphics[width=0.8\linewidth]{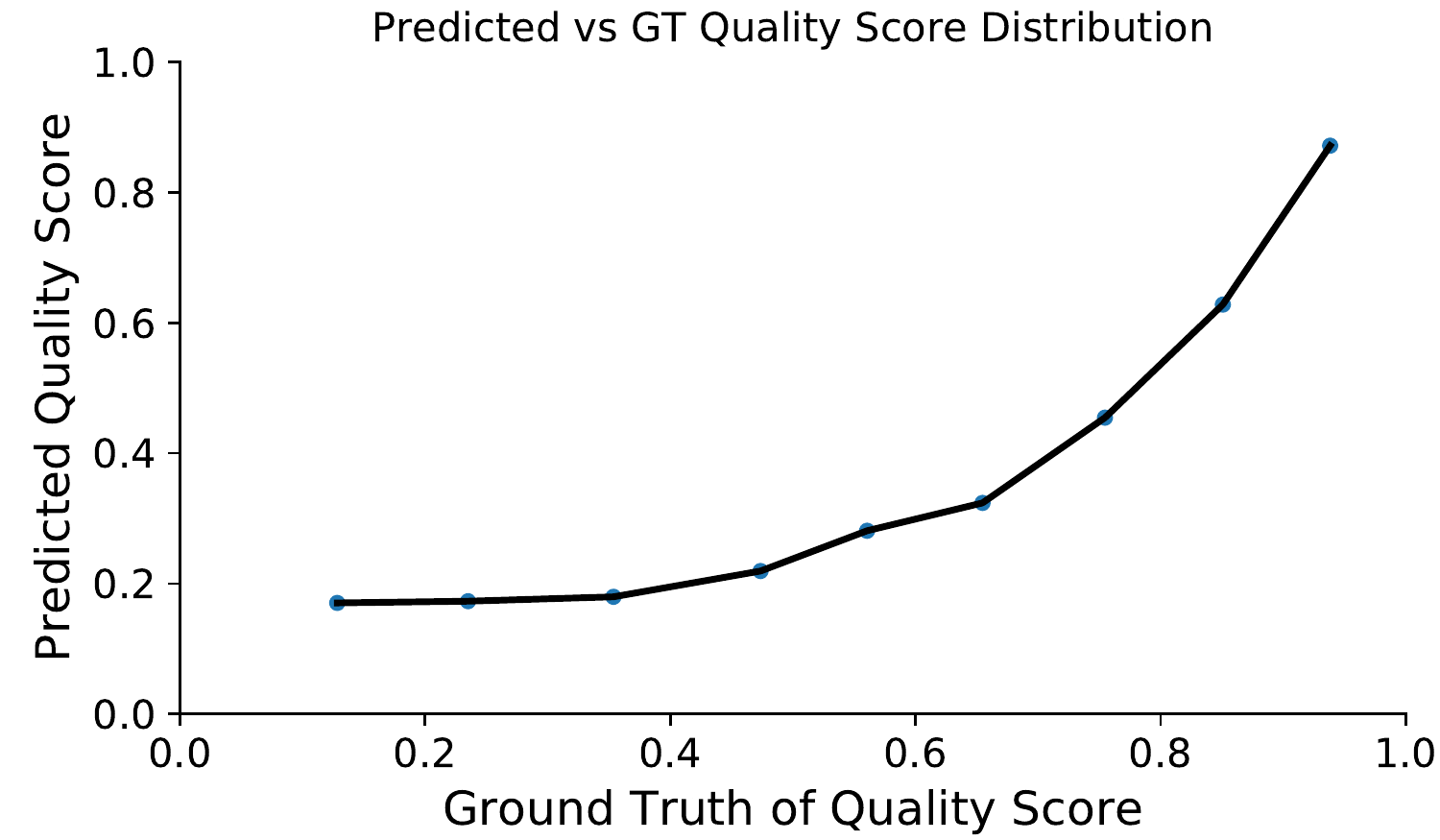}}
\end{center}
\vspace{-0.5cm}
\caption{The quality distribution curves on different scenarios. (a) The predicted quality score curves under different interferences (`Blur', `Illumination' and `Mixed') on synthetic MNIST. (b) The predicted quality score curve on CCPD.}
\label{fig_score_curve}
\end{figure}

\textbf{Overall performance}.
Table \ref{tab_mnist} shows that our method achieves promising quality assessment results under all conditions.
And Figure \ref{fig_score_curve_mnist} implies the predicted scores also appear in approximate distribution with their ground truth under different interferences, which verifies the robustness of our method.

\begin{figure}
\begin{center}
\includegraphics[width=0.99\linewidth]{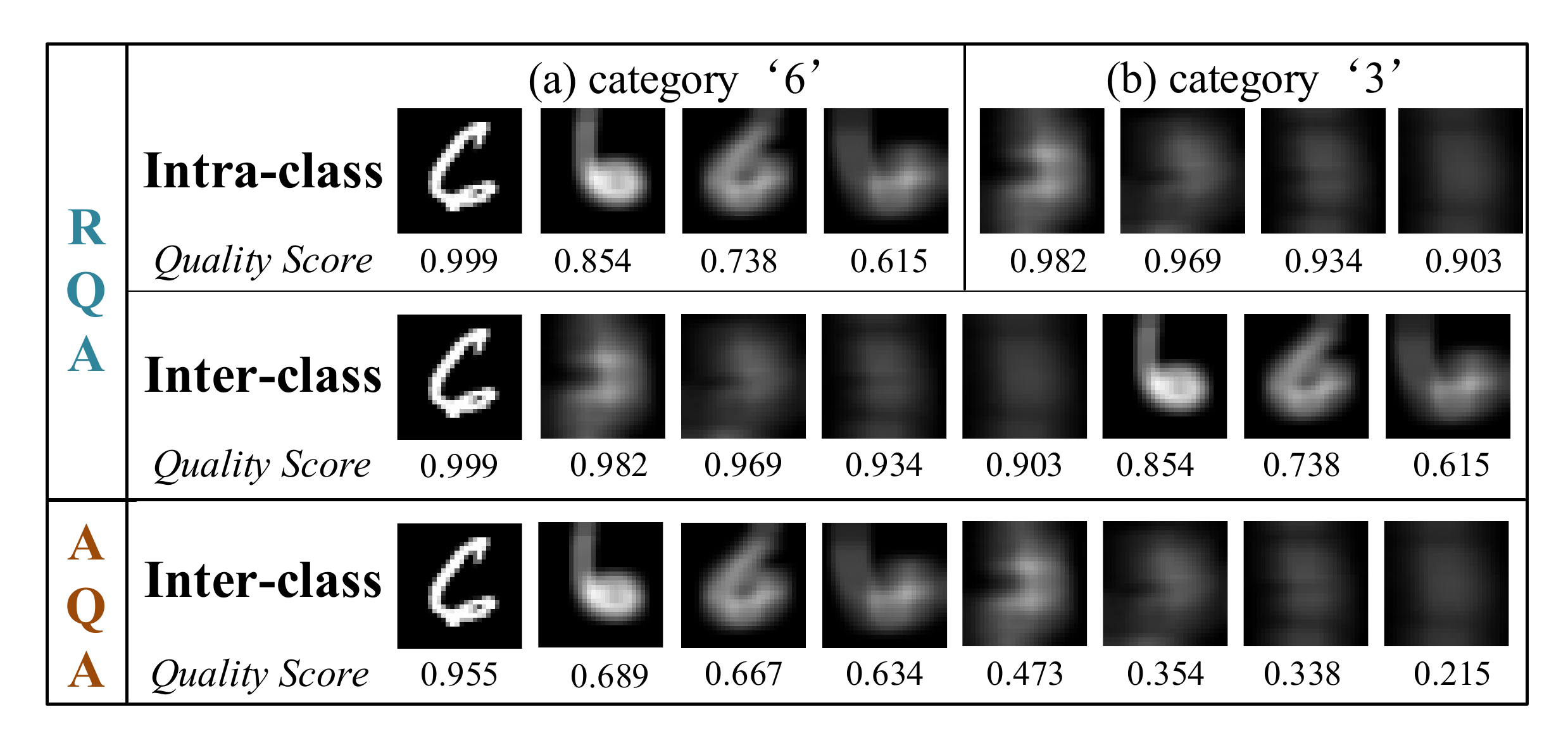}
\vspace{-0.5cm}
\end{center}
   \caption{The visualization of quality scores by RQA and AQA. 
   }
\label{fig_visualize_mnist}
\end{figure}

\textbf{Effects of RQA and AQA}.
In Table \ref{tab_mnist}, we observed that both RQA and AQA succeed in assessing quality scores in intra-class, and AQA significantly outperforms RQA on `Inter' sets under all conditions (See `Inter' rows).
An example is illustrated in Figure \ref{fig_visualize_mnist}, the intra-class quality ranking of RQA seems rational but it fails on assessing quality scores in inter-class,
while AQA can generate satisfying quality ranking in inter-class.

\begin{table}
\begin{center}
\scalebox{1}{
\begin{tabular}{lcccc}
\hline
$\mathcal{L}_{align}$ & \checkmark   & \checkmark       & \checkmark   & \checkmark  \\
$\mathcal{L}_{intra}$ &              & \checkmark       &              & \checkmark  \\
$\mathcal{L}_{inter}$ &              &                  & \checkmark   & \checkmark  \\
\hline
SROCC  & 0.774 & 0.792 & 0.802 & \textbf{0.814} \\
\hline
LCC  & 0.832 & 0.863 & 0.869 & \textbf{0.895} \\
\hline
\end{tabular}
}
\end{center}
\caption{
    The evaluation of AQA's constraints on the `Mixed' set.
}
\vspace{-0.5cm}
\label{tab_aqa_ablation}
\end{table}

\textbf{Exploration on AQA}.
We further validate the effect of each constraint in AQA, as shown in Table \ref{tab_aqa_ablation}.
Since $\mathcal{L}_{align}$ is used to provide the quality anchor for aligning scores between different classes, it is indispensable in our method.
We find that both $\mathcal{L}_{intra}$ and $\mathcal{L}_{inter}$ can significantly improve the results of SROCC and LCC, and the jointly learning further improves the performance. 

\textbf{Inference speed}. AQA is very light-weighted and its inference speed achieves 1820 frames per second as we test on 12GB Titan-X GPU with `PyTorch'.

\begin{figure*}
\begin{center}
\includegraphics[width=0.99\linewidth]{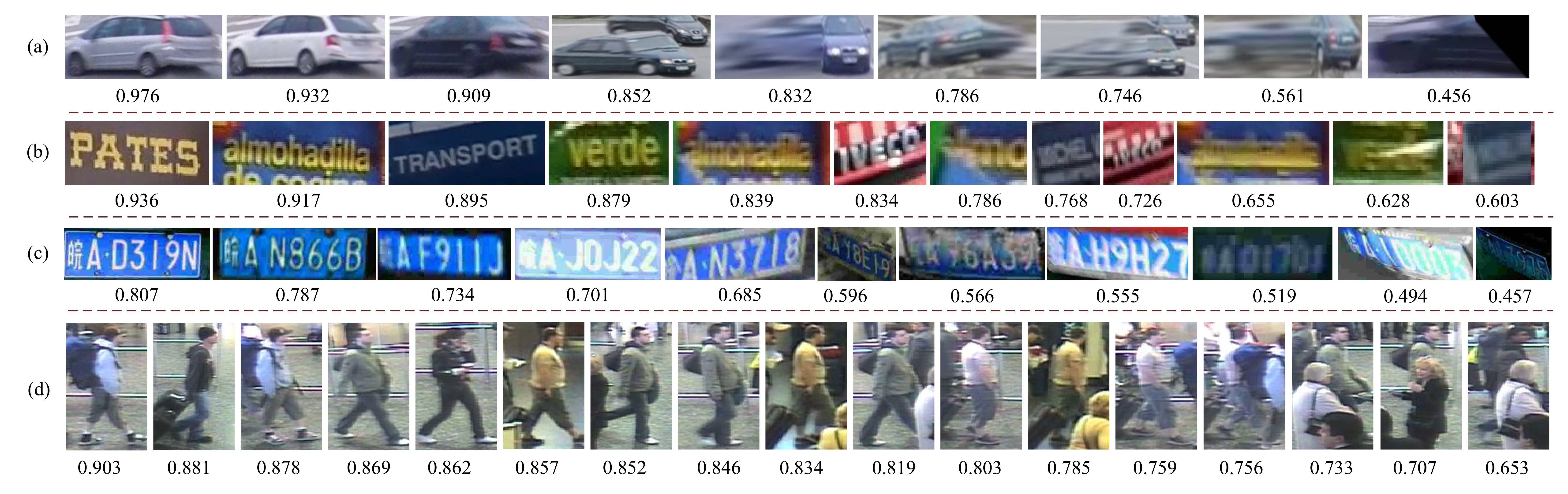}
\end{center}
\vspace{-0.5cm}
   \caption{The quality scores (under each image) generated by AQA on 4 tasks. (a), (b), (c) and (d) show the predicted quality scores in vehicle re-identification, video text spotting, license plate recognition and person re-identification scenario, respectively.
   }
\label{fig_visualize_4tasks}
\end{figure*}

\subsection{Evaluation on Vehicle Re-identification}
\textbf{Dataset}.
We choose VehicleReId dataset \cite{Zapletal2016Vehicle} for our evaluation on vehicle re-identification task. 
Half of the dataset is used as training set while the rest as testing set.
We adopt the model used in \cite{Liu2018PROVID} as pre-trained recognition model, and the outputs before final classification layer are used as the extracted features.

\begin{table}
\begin{center}
\scalebox{1}{
\begin{tabular}{ccc|cc|cc}
\hline
\multicolumn{1}{c|}{Dataset} & \multicolumn{2}{c|}{VehicleReId} & \multicolumn{2}{c|}{IC13} & \multicolumn{2}{c}{CCPD} \\
\hline
\multicolumn{1}{c|}{Metric} & $SRC_{a}$ & $SRC_{e}$ & $SRC_{e}$ & $QSHR$ & $SRC_{a}$ & $SRC_{e}$ \\
\hline
\multicolumn{1}{c|}{PCW \cite{Cheng2019You}} & 0.736 & 0.641 & 0.517 & 0.745 & 0.565 & 0.693 \\
\multicolumn{1}{c|}{QAN \cite{liu2017quality}} & 0.783 & 0.675 & 0.716 & 0.809 & - & - \\
\multicolumn{1}{c|}{YORO \cite{Cheng2019You}} & 0.765 & 0.709 & 0.733 & 0.817 & - & - \\
\hline
\multicolumn{1}{c|}{RQA} & \textbf{0.846} & 0.748 & 0.748 & 0.821 & \textbf{0.697} & 0.758\\
\multicolumn{1}{c|}{AQA} & 0.837 & \textbf{0.853} & \textbf{0.761} & \textbf{0.827} & 0.651 & \textbf{0.855} \\
\hline
\end{tabular}
}
\end{center}
\caption{Quantitative evaluations of quality metrics on VehicleReId, IC13 and CCPD referring to human annotations, including comparison with related existing approaches. $SRC_{a}$ and $SRC_{e}$ are `intra-class' and `inter-class' SROCC respectively, all the existing approaches are re-implemented with our backbone.
}
\vspace{-0.5cm}
\label{tab_srocc}
\end{table}

\textbf{Quality assessment result}.
We visualize quality scores in Figure \ref{fig_visualize_4tasks} (a). which shows that vehicle images of different qualities from different classes can be well distinguished.
For quantitative evaluations on quality scores, we randomly select 48 groups of images from the testing set (each group contains 3 different quality images), in which 24 groups contain images from the same class while the rest contain images from different classes.
Then we equally divide them into 6 subsets (each subset contains 4 groups of intra-class images and 4 groups of inter-class images). Each subset is annotated by 1 person to decide
the quality ranking in intra-class and inter-class, which is regarded as ground truth.
Finally, we compute average SROCC on these subsets, shown in Table \ref{tab_srocc}, $SRC_{a}$ and $SRC_{e}$ both indicate that our method outperforms all the existing strategies \cite{liu2017quality,Cheng2019You} and achieves more consistent quality assessment with human perception, 
especially the $SRC_{e}$ of AQA is significantly higher (0.15) than state-of-art strategies.

\begin{table}
\begin{center}
\scalebox{1}{
\begin{tabular}{cccc|ccc}
\hline
\multicolumn{1}{c|}{Dataset} & \multicolumn{3}{c|}{VehicleReId} & \multicolumn{3}{c}{iLIDS-VID} \\
\hline
\multicolumn{1}{c|}{Metric} & R1 & R5 & R10 & R1 & R5 & R10 \\
\hline
\multicolumn{1}{c|}{BASE} & 75.1\% & 92.9\% & 95.8\% & 69.3\% & 90.7\% & 94.0\% \\
\multicolumn{1}{c|}{QAN \cite{liu2017quality}} & 75.5\% & 92.1\% & 95.8\% & 71.2\% & 89.6\% & 94.0\% \\
\multicolumn{1}{c|}{YORO \cite{Cheng2019You}} & 75.8\% & 92.5\% & 95.8\% & 71.5\% & 90.2\% & 94.0\% \\
\hline
\multicolumn{1}{c|}{BASE+AQA} & \textbf{77.1\%} & 92.9\% & 95.8\% & \textbf{73.3\%} & 90.7\% & 94.0\% \\
\hline
\end{tabular}
}
\end{center}
\caption{The cumulative matching characteristics on VehicleReId and iLIDS-VID. `R1' denotes rank-1 result.
}
\vspace{-0.5cm}
\label{tab_reid}
\end{table}

\textbf{Performance improvement}.
Table \ref{tab_reid} shows the performance gain.
Following \cite{GaoRevisiting}, we use temporal average pooled feature of each vehicle sequence as its representation, and report its results as baseline (denoted as `BASE').
Then we discard images with scores lower than the learned threshold (\ie the learnable quality anchor $th$ (0.839) in Equation \ref{align_loss}) in each sequence to obtain its new feature representation, and report the results as `BASE+AQA'.
By filtering out low-quality images, `BASE+AQA' achieves 2\% gain at rank-1 (R1).

\subsection{Evaluation on Video Text Spotting}
\textbf{Dataset}.
The ICDAR 2013 `Text in Video' dataset (\emph{abbr.} IC13) \cite{karatzas2013icdar} is used for our evaluation on video text spotting.
IC13 provides coarse quality annotations, i.e. each image is labeled with `HIGH', `MODERATE' or `LOW' quality level.
The model used in \cite{cheng2017focusing} is treated as our pre-trained recognition model, and the `glimpse' features in its attention decoder is used as the extracted features.

\textbf{Quality assessment result}.
We first visualize the quality scoring results in Figure \ref{fig_visualize_4tasks} (b), in which the predicted quality scores are also satisfying.
For quantitative evaluations,
we calculate the average quality score of `HIGH', `MODERATE' and `LOW' images in each text sequence, denoted by $(\overline{s_h},\overline{s_m},\overline{s_l})$.
Apparently, the ground truth quality ranking should be $\overline{s_h} > \overline{s_m} > \overline{s_l}$, then SROCC metric can be computed by comparing the predicted ranking with ground truth.
Besides, we also adopt `QSHR' used in \cite{Cheng2019You} for evaluating the accuracy of selecting the highest-quality image in each sequence.
The results are shown in Table \ref{tab_srocc}, compared with PCW \cite{Cheng2019You} (using recognition confidence as quality score)
, QAN \cite{liu2017quality} and YORO \cite{Cheng2019You}, our method outperforms them by at least 1\%. 

Additionally, we separately compute the average score of all the `HIGH', `MODERATE' and `LOW' images on the entire testing set, which are 0.661, 0.606 and 0.394. Notice that the estimated average score difference (0.055) between `HIGH' and `MODERATE' is far less than the difference (0.212) between `LOW' and `MODERATE'.
Thus we randomly selected 15 high-quality image pairs (1 `HIGH' image and 1 `MODERATE' image in each pair) and 15 low-quality image pairs (1 `MODERATE' image and 1 `LOW’ image in each pair), then ask 2 persons to annotate the quality ranking in each pair.
Consequently, their decisions are consistent on all the low-quality image pairs, but inconsistent on 4 high-quality image pairs, which refects that the quality difference between `HIGH' and `MODERATE' images is harder to distinguish compared to that between `MODERATE' and `LOW' images.

\begin{table}
\begin{center}
\scalebox{1}{
\begin{tabular}{cc|c}
\hline
\multicolumn{1}{c|}{Dataset} & \multicolumn{1}{c|}{IC13} & \multicolumn{1}{c}{CCPD} \\
\hline
\multicolumn{1}{c|}{Metric} & SRA & SRA \\
\hline
\multicolumn{1}{c|}{BASE} & 60.72\% & 82.30\% \\
\multicolumn{1}{c|}{QAN \cite{liu2017quality}} & 60.99\% & - \\
\multicolumn{1}{c|}{YORO \cite{Cheng2019You}}& 61.45\% & - \\
\hline
\multicolumn{1}{c|}{BASE+AQA} & \textbf{62.99\%} & \textbf{88.61\%} \\
\hline
\end{tabular}
}
\end{center}
\caption{The \textbf{s}equence-level \textbf{r}ecognition \textbf{a}ccuracy (SRA) improvements on IC13 and CCPD.}
\vspace{-0.5cm}
\label{tab_text_recog}
\end{table}

\textbf{Performance improvement}.
Table \ref{tab_text_recog} shows the recognition accuracy gain by adopting our quality scores.
The baseline (denote by `BASE') selects the most frequently occurred recognition results as the final result for each sequence, which is similar to majority voting \cite{wang2017end}.
While equipped with our method (denoted by `BASE+AQA'), we discard images with scores lower than the learned threshold (0.711) by AQA, then proceed the voting strategy to get final recognition results.
We find that our quality assessment can help recognition model improve accuracy by 2.27\%, largely surpassing the 0.73\% of YORO \cite{Cheng2019You}.

\subsection{Evaluation on License Plate Recognition}
\textbf{Dataset}.
The large-scale license plate dataset `CCPD' \cite{Xu2018Towards} is selected for quality assessment evaluation.
It contains over 300 thousands of plates, in which many images are polluted by various noises and detailed annotations of brightness and blurriness are available.
Here, 66\% images are randomly selected as our training set and the rest as testing set.
The recognition model architecture is the same as that on IC13, and pre-trained on CCPD.

\textbf{Quality assessment result}.
We visualize the generated quality scores in Figure \ref{fig_visualize_4tasks} (c), and the predictions are also satisfying.
For quantifying the quality assessment performance, we construct a sub test set (containing 508 images) by selecting images with various brightness and blurriness.
Then we generate their ground truth of quality scores by two steps:
1) since all plates follow the same character layout (\eg the position of each character in a standard license plate is fixed), we synthesize the standard plate images with no interference as their standard templates; 
2) following the same strategy on MNIST, cosine similarity between the feature of testing image and its standard template is calculated as the ground truth, which is further normalized to the range [0,1].
SROCC on this sub set are shown in Table \ref{tab_srocc}, our method outperforms the naive strategy PCW \cite{Cheng2019You} by a large margin (0.13/0.16 on $SRC_{a}$/$SRC_{e}$), and
the $SRC_{e}$ of AQA is significantly higher than that of RQA, which indicates AQA can produce much more reliable absolute quality scores in inter-class. We haven't re-implement other existing strategies (QAN \cite{liu2017quality} and YORO \cite{Cheng2019You}) on this dataset since they all require video frames as input, which can't be satisfied in this scenario.

Furthermore, we divide score range $[0,1]$ into 10 equal intervals $[0,0.1]$, $...$, $[0.9,1]$, then generate 10 subsets according to the ground truth score of each image. 
We compute the predicted average score of each subset, and analyze its distribution in
Figure \ref{fig_ccpd_qscore_distri}, which implies that our predicted scores remain consistent with ground truth whether images are polluted by blur or brightness.

\textbf{Performance improvement}.
We select 729 groups of license plates from the total test set for evaluating the recognition performance gain, each group contains at least 3 images, and experiment setting is the same as that on IC13. 
The performance improvement is illustrated in Table \ref{tab_text_recog}, which shows the recognition accuracy is significantly improved by 6.3\% by discarding low-quality images.

\subsection{Evaluation on Person Re-identification}
\textbf{Dataset}.
We choose iLIDS-VID dataset \cite{Taiqing2014Person} for evaluations on person re-identification, because it contains more diversified quality images compared with other person re-identification datasets.
We obtain the pre-train model by following \cite{GaoRevisiting}, then the outputs before the final classification layer are used as the extracted features.

\textbf{Quality assessment result}.
Due to lack of ground truth quality scores, we only visualize and observe whether predicted scores are consistent with human perception in this scenario, shown in Figure  \ref{fig_visualize_4tasks} (d).
As a whole, predicted scores can reasonably express the absolute quality relationships among different class person images.


\textbf{Performance improvement}.
We further evaluate the power of quality assessment on improving person re-identification performance. As shown in Table \ref{tab_reid}, `BASE' and `BASE+AQA' share the same evaluation settings as that on vehicle re-identification,
our method significantly improves the top-1 matching (R1) by 4\%, outperforming the improvements brought by existing strategies \cite{liu2017quality,Cheng2019You}, which proves the advantage of our method compared with state-of-art.
\section{Conclusion}
In this paper, we define the new problem of \emph{object quality assessment} and propose an object quality assessment approach named Object-QA to assess high-reliable quality scores, which can help improve the performance and time efficiency of various recognition applications.
Inspired by human cognition, Object-QA is implemented by two progressive modules: RQA for obtaining intra-class-level quality scores and AQA for generating inter-class-level quality scores.
Extensive experiments on four real-life tasks demonstrate that Object-QA can provide high-reliable quality scores which not only accords with human cognition and also improves recognition performance.
In future, we hope quantitative benchmarks can be established and we’ll further explore the object quality assessment in more diversified scenarios.


\bibliographystyle{ACM-Reference-Format}
\bibliography{egbib}

\end{document}